\documentclass[letterpaper, 10 pt, conference]{ieeeconf}  

\IEEEoverridecommandlockouts  
\usepackage[pdftex]{graphicx}
\DeclareGraphicsExtensions{.png,.pdf}
\usepackage{amsmath} 
\usepackage{amssymb}  
\DeclareMathOperator{\E}{\mathbb{E}}
\DeclareMathOperator{\D}{\mathbb{D}}

\usepackage{tabularx,booktabs,multirow}
\usepackage{array}
\usepackage{float}
\usepackage{makecell}
\usepackage{tablefootnote}
\usepackage{multicol}
\usepackage{booktabs}
\usepackage{threeparttable}
\usepackage{url}
\usepackage[font=small,skip=10pt]{caption}
\usepackage[dvipsnames]{xcolor}
\usepackage{amsmath}
\usepackage{xcolor}
\usepackage{algorithmic}
\usepackage{dirtytalk}
\usepackage[ruled,vlined]{algorithm2e}

\newcolumntype{P}[1]{>{\centering\arraybackslash}p{#1}}
\SetKwInOut{Input}{input}
\SetKwInOut{Output}{output}
\makeatletter
\let\NAT@parse\undefined
\makeatother
\usepackage{hyperref}
\hypersetup{colorlinks,allcolors=red}

\title{\LARGE \bf
Fusing Visuo-Tactile Perception into Kernelized Synergies for Robust Grasping and Fine Manipulation of Non-rigid Objects 
}

\author{Sunny Katyara$^{1,2}$, ~\IEEEmembership{Student Member,~IEEE}, Nikhil Deshpande$^{1}$,  ~\IEEEmembership{Member,~IEEE}, Fanny Ficuciello$^{2}$,\\ ~\IEEEmembership{Senior Member,~IEEE}, Fei Chen$^{3}$,~\IEEEmembership{Senior Member,~IEEE},
Bruno Siciliano$^{2}$,~\IEEEmembership{Fellow,~IEEE}, \\ Darwin G. Caldwell$^{1}$,~\IEEEmembership{Senior Member,~IEEE} 
\thanks{This research is supported by the projects – “LEARN-REAL” funded by EU H2020 ERA-Net Chist-Era; “HARMONY” funded by EU H2020 R\&I under agreement No. 101017008; and by the Italian Workers’ Compensation Authority (INAIL) under the “Sistemi Cibernetici Collaborativi” agreement.
(INAIL). \textit{(Corresponding author: Darwin G. Caldwell)} }
\thanks{$^{1}$ Department of Advanced Robotics, Istituto Italiano di Tecnologia, Via Morego 30, 16163, Genova, Italy (e-mail: {\tt\small name.surname@iit.it}).}
\thanks{$^{2}$ Department of Information Technology and Electrical Engineering and PRISMA Lab, University of Naples Federico II, Naples 80125, Italy ({e-mail: \tt\small name.surname@unina.it}).}
\thanks{$^{3}$ Department of Mechanical and Automation Engineering, The Chinese University of Hong Kong, Hong Kong ({e-mail: \tt\small f.chen@ieee.org}).}
}

\begin{document}

\maketitle
\thispagestyle{empty}
\pagestyle{empty}

\begin{abstract}

 Handling non-rigid objects using robot hands necessities a framework that does not only incorporate human-level dexterity and cognition but also the multi-sensory information and system dynamics for robust and fine interactions. In this research, our previously developed kernelized synergies framework, inspired from human behaviour on reusing same subspace for grasping and manipulation, is augmented with visuo-tactile perception for autonomous and flexible adaptation to unknown objects. To detect objects and estimate their poses, a simplified visual pipeline using RANSAC algorithm with Euclidean clustering and SVM classifier is exploited. To modulate interaction efforts while grasping and manipulating non-rigid objects, the tactile feedback using T40S shokac chip sensor, generating 3D force information, is incorporated. Moreover, different kernel functions are examined in the kernelized synergies framework, to evaluate its performance and potential against task reproducibility, execution, generalization and synergistic re-usability. Experiments performed with robot arm-hand system validates the capability and usability of upgraded framework on stably grasping and dexterously manipulating the non-rigid objects.  

\end{abstract}

\section{INTRODUCTION} Prehensile robotic manipulation requires to infer the states of system and objects in the environment for efficient task planning and control \cite{c1}\cite{c2}. The task planning not only relies upon proprioception but also exteroception i.e, \textbf{visuo-tactile information} to effectively perceive the environment i.e, nature and placement of objects in the world. The robots having higher degrees of freedom (DOF) i.e, artificial hands, raises concern over its controller complexity and thus necessitates to exploit reduced subspace such as postural synergies \cite{c3} \cite{c4} for robust grasping and fine manipulation. The \textbf{Robust grasping} refers to the ability of robot hand to maintain the grasp stability against external disturbances exerted on the grasped object whereas the \textbf{fine manipulation} involves moving an object to a desired position primarily with the fingertips \cite{c5}. Both the actions need a fine regulation of interaction forces to avoid slippage and damage to the non-rigid objects or the robot itself.  

   \begin{figure}[t]
      \centering
      \includegraphics[width=8.5cm]{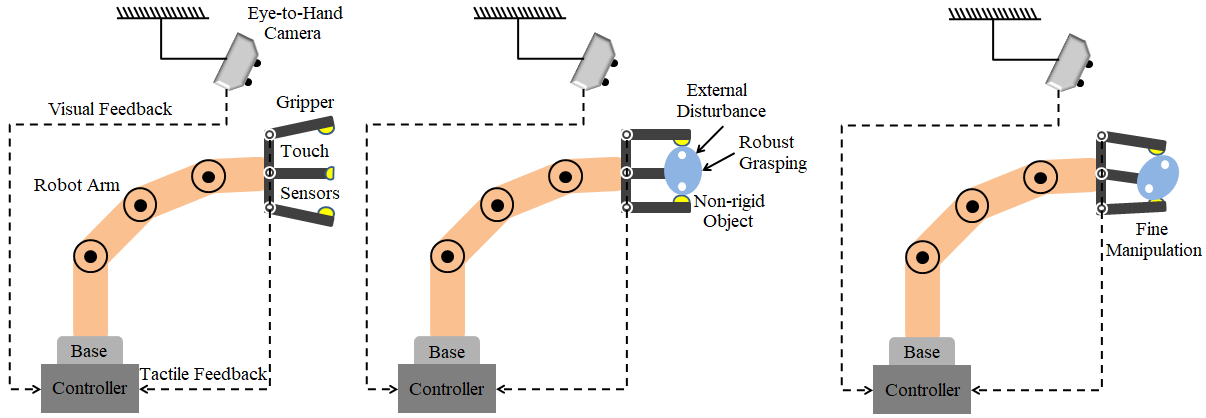}
      \caption{Sketch of visuo-tactile adaptation to robust grasping and fine manipulation of non-rigid object. Camera monitors the object in the scene and helps in enumerating its pose for grasp formation. Touch sensors modulate the force profile of the gripper for flexible interaction during the grasping and manipulation actions.}
      \label{representation}
      \vspace{-15pt}
   \end{figure}

Following the idea of postural synergies, a framework called \textbf{kernelized synergies} has been developed that retains the uniform subspace for precision grasping and in-hand manipulation \cite{c6}. The framework however is able to deal with the new objects for task planning and execution but the information about its environment is encoded manually. Therefore, to automate the process of object detection, pose estimation and force adaptation at run-time, it requires to incorporate the visuo-tactile perception into the framework, as shown in Fig. \ref{representation}. In general, the visuo-tactile based robotic manipulation involves three key steps; \textbf{(1)} localizing and estimating the pose of candidate objects, \textbf{(2)} planning and executing the desired grasp, and \textbf{(3)} modulating the force profile during robot-object interaction.   

Exploiting visuo-tactile perception into robot-object interaction tasks entails to implement one of the three control strategies i.e, traded, hybrid and shared \cite{c7}. In traded control, the visual and tactile information are exploited consecutively, based on the pre-defined task threshold. However, the hybrid scheme fuses both the sensory feedback but exploit them independently in an orthogonal space, according to the description of their task frames. Whereas, in a shared architecture, both the sensory feedback are co-utilized concurrently in the common subspace. In \cite{c8}, a shared scheme combing visual and haptic information for precise grasping of simple rigid objects is presented. The proposed idea is a standard approach to fuse distinct sensory data for robotic grasping but is prone to variations in the environment and system dynamics. A hybrid framework, being robust against model and system uncertainties, is proposed in \cite{c9} for stable grasping and dexterous manipulation of non-rigid objects. This method works well in the structured environment and on the known objects but it is a object-centric approach and suffers from controller complexity. To alleviate the curse of controller complexity and model bias, traded strategies employing the idea of synergistic control, are discussed in \cite{c10}\cite{c11}. These techniques effectively deal with the perceptual uncertainties and dynamic variations but do not take into account the interaction flexibility and are also limited to the manipulation of known rigid objects. Therefore, in this research, we propose to exploit visuo-tactile perception in a traded fashion to introduce run-time adaptation and interaction flexibility apart from synergistic reusability and grasp stability into the kernelized synergies framework \cite{Katyara2021}.     

With reference to \cite{c6}\cite{c11}, the major contributions of this research are; \textbf{(1)} to augment the kernelized synergies framework with visuo-tactile feedback for autonomous and flexible object manipulation, \textbf{(2)} to investigate the performance and potential of kernelized synergies using different kernels i.e, Exponential, Gaussian and Cauchy against the subspace re-usability and task reproducibility, generalization, and execution, \textbf{(3)} to test the updated framework to perform two distinct tasks including; picking and placing egg into the placeholder and pouring tomato ketchup onto the food.   

\section{RESEARCH METHODOLOGY}

   \begin{figure}[t]
      \centering
      \includegraphics[width=8.5cm]{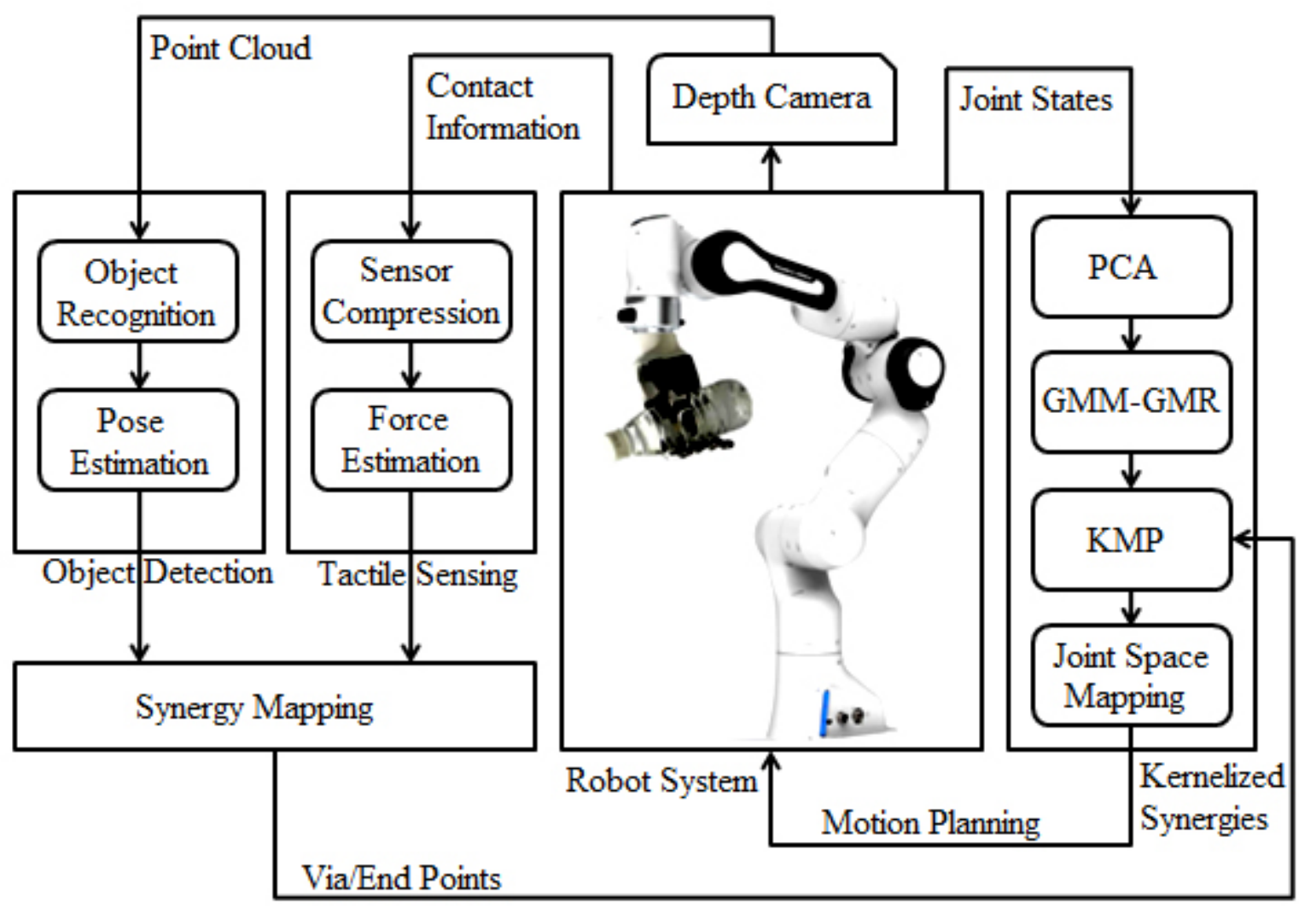}
      \caption{Kernelized synergies framework with visuo-tactile feedback.}
      \label{methodology}
      \vspace{-15pt}
   \end{figure}

Figure \ref{methodology} shows the block diagram of updated kernelized synergies framework augmented with visuo-tactile feedback for anthropomorphic grasping and manipulation of non-rigid objects. The framework uses a dimensionality reduction technique i.e, PCA (Principal Component Analysis) to extract predominant postural synergies from the robot hand joint configurations. The co-efficients of the computed synergies are interpolated over the duration of demonstrations to obtain the corresponding synergistic trajectories. In order to generalize the subspace to the given set of demonstrations and to reproduce the learned tasks, the computed synergistic trajectories are encoded in terms of Gaussian components and a reference synergistic trajectory is generated using GMM-GMR (Gaussian Mixture Model - Gaussian Mixture Regression). To reuse the computed synergistic subspace for grasping and manipulation of unknown objects, the probabilistic synergistic trajectories are parametrized and kernelized using KMP (Kernelized Movement Primitives). The kernel trick used in KMP helps to preserve the interaction properties of the synergistic subspace globally so that it can be reused for interacting with distinct new objects using its environmental descriptors i.e, via and end-points for the shape and size respectively. Therefore, to estimate the geometrical features of the candidate objects, the visual information in terms of point cloud using RANSAC algorithm together with the SVM classifier updates the environmental descriptors of the kernelized synergies framework. With an object being grasped and manipulated in the kernelized synergistic subspace, the force profile of the robot hand get updated with the tactile feedback. Both the sensory data are transformed into synergistic values for homogeneous modulation to the reference trajectory in the framework. Finally, a mapping to joint space from synergistic subspace enables to execute the defined tasks on the robot system.

\subsection{Kernelized Synergies Framework}

\begin{figure*}[t]
      \centering
      \includegraphics[height=3.5cm, width=17cm]{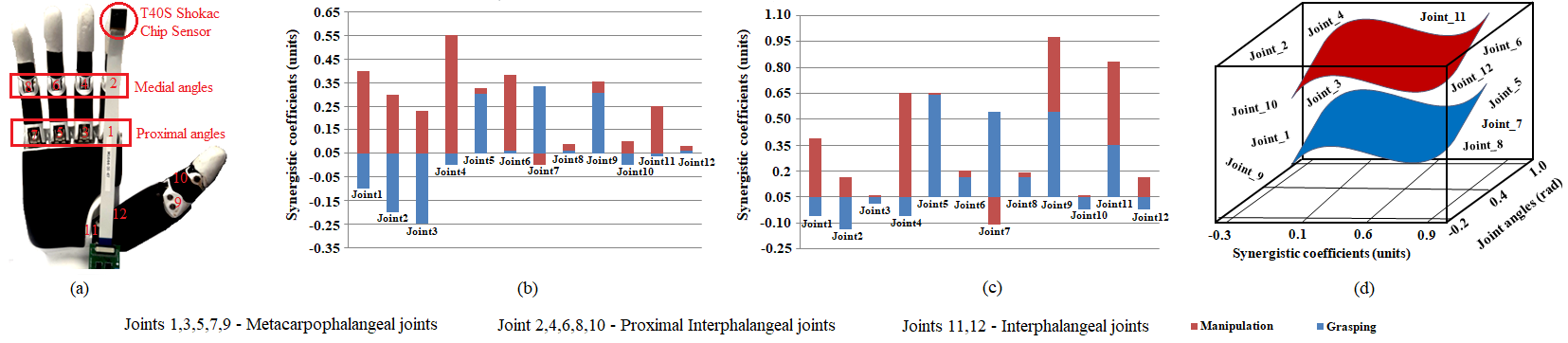}
      \caption{Graphical representation of the computed synergistic subspace of the anthropomorphic robot hand, (a) represents the 6 DOF robot hand with a tactile sensor installed at its fingertip, (b) and (c) are the first two predominant synergies with respective grasping $(e_g)$ and manipulation $(e_m)$ components, (d) is the kernelized synergistic subspace of two predominant synergistic components.}
      \label{synergies}
      \vspace{-15pt}
   \end{figure*}

The robot hand shown in Fig. \ref{synergies} \cite{c12}, is tele-operated and the mean hand joint configurations ${\hat{\vartheta_k}}$=$\vartheta_k$-$\theta_0$ are recorded, where $\vartheta$ and $\theta$ represent the current and nominal hand postures respectively. The mean hand joint configurations for each demonstration are then concatenated into a row vector to build an array, called configuration matrix $C$=$[{\hat{\vartheta_1}}......{\hat{\vartheta_K}}]^{T}$. The PCA is applied to $C$ to determine its lower dimensional representation i.e, synergistic subspace, which is numerically characterized by the synergy matrix $\hat{E}$. With a proper choice of coefficients i.e, $e_i={e_{g1}+e_{m1},e_{g2}+e_{m2},...e_{gn}+e_{mn}}$, the synergistic subspace can be properly accessed for grasping (g) and manipulation (m) of candidate objects. Their values are enumerated using Eq. \ref{eqt_1}, where ${\hat{E}}^{\dagger}$ is the pseudo inverse of the synergy matrix.

\begin{equation}
{e_i}={\hat{E}}^{\dagger}{({\vartheta_i}-{\theta})}
\label{eqt_1}
\end{equation}

The subspace of the robot hand, consisted of two predominant synergistic components (i.e, having variance greater than or equal to $85\%$) in Figs. \ref{synergies} (b and c), is shown in Fig. \ref{synergies} (d). It is evident from Figs. \ref{synergies} (b and c) that the first synergistic component $(e_{g1}+e_{m1})$ primarily controls the proximal and medial angles of all the fingers whereas the second component $(e_{g2}+e_{m2})$ regulates the relative movements of thumb and index finger, which help to switch from one posture to other during object manipulation. Fig. \ref{synergies} (d) reveals that the two components are however numerically independent but are synergistically correlated for task imitation and execution during model inference.

The coefficients of the synergy matrix in Eq. \ref{eqt_1} are interpolated over the duration of object demonstration $(t)$ to determine their corresponding synergistic trajectories $e(t)$. In order to account for inconsistencies in the demonstrations and also to deal with the feedback uncertainities, the resultant synergistic trajectories are approximated in terms of Gaussian's using GMM i.e, $e(t){\sim}{\sum_{n=1}^{N}}{\pi_n}{\mathcal{N}}({{\mu}_n},{{\Sigma}_{n}})$, where ${\pi}_n$,${\mu}_n,{\Sigma}_n$ represent the prior, mean and co-variance of $n^{th}$ Gaussian respectively and N is the total number of Gaussian components used. Further, to reproduce the learned hand joint configurations, a reference trajectory is generated by using GMR that conditions the joint probability distribution from the GMM i.e, ${e_n}|t{\sim}{\mathcal{N}}({{\mu}_n},{{\Sigma}_n})$. 

To generalize and reuse the computed synergistic subspace over the wider set of objects, the reference synergistic trajectory is parameterized and kernelized using KMP. For an instance of a new object $t^{*}$, the expected mean and co-variance of the reference synergistic trajectory are determined by using Eq. \ref{eqt_2} \cite{c13}, where $k^{*}$ and $K$ are the kernel function and matrix respectively and $\lambda$ is a regularization parameter, that helps to avoid variance in the model predictions. 

\begin{equation}
\begin{gathered}
{\E}(e(t^{*}))={k}^{*}({K}+{\lambda}{I})^{-1}{\mu} \\
\D(e(t^{*}))={\frac{N}{\lambda}}({k}({t}^{*},{t}^{*})-{k}^{*}({K}+{\lambda}{\Sigma})^{-1}k^{*T})
\end{gathered}
\label{eqt_2}
\end{equation}

The kernelized synergies can also prioritize different synergistic movements thereby assigning appropriate weight/priority coefficient $(\Upsilon)$ to the synergistic components and is thus defined by Eq. \ref{eqt_3} \cite{c13}, which happens to be the product of $D$ Gaussian distributions approximating the given reference synergistic trajectory, with $d= 1,2,...D$.

\begin{equation}
\mathcal{N}({\mu_n}^{T},{\Sigma_n}^{T}) {\propto}{\prod_{d=1}^{D}}\mathcal{N}({\hat{\mu}_{n,d}},{\hat{\Sigma}_{n,d}}/\Upsilon_{n,d})
\label{eqt_3}
\end{equation}

\subsection{Object Segmentation and Localization}

\begin{figure*}[t]
      \centering
      \includegraphics[height=4.4cm, width=17.6cm]{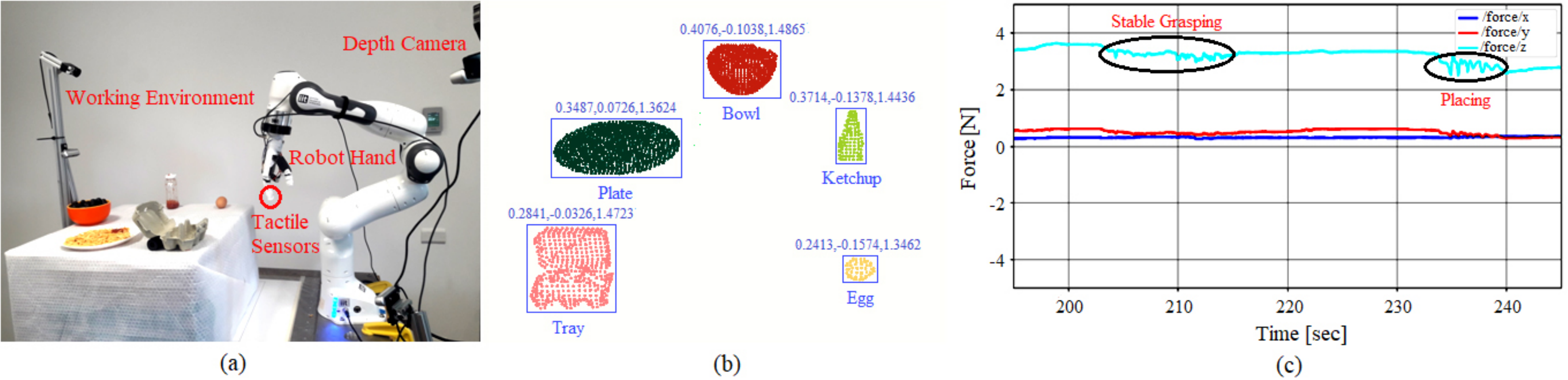}
      \caption{Proposed setup for grasping and manipulation of non-rigid objects using visuo-tactile perception, (a) represents the experimental environment for autonomous and flexible robot-object interaction, (b) illustrates the point cloud of the segmented and classified candidate objects in the scene, (c) is the force feedback on grasping and placing of a non-rigid object (grape) into the bowl.}
      \label{pose}
      \vspace{-15pt}
   \end{figure*}

The object instances $t^{*}$ (i.e, the duration of geometrical features of the objects in the synergistic subspace) in Eq. \ref{eqt_2}, are determined from the point cloud of the scene that is acquired using depth camera, as shown in Fig. \ref{pose} (b). The point cloud is processed using RANSAC algorithm which separates out the candidate objects from the background in terms of outliers and inliers. In order to group the relevant points of the candidate objects, the Euclidean clustering criteria, defined by Eq. \ref{eqt_5}, is applied.         

\begin{equation}
    O = {\int_{j=0}^{K_C}}N_{\epsilon}(A_j|M_{np}):{\{B_j|d(A_j,B_j){\leq}{\epsilon}\}}dA
\label{eqt_5}
\end{equation}

\begin{algorithm}
\SetAlgoLined
\Input{${O_p, P_c, q_i}\longrightarrow{\text{poses, contacts, hand joints}}$}   
\Output{${\rho}(e(t))\longrightarrow{\text{synergistic reference trajectory}}$}
 \While{$({q_{i}}\in{q^{k}})\longrightarrow{C}$}{
 $function (C)\longrightarrow(\hat{E},e)$\\
 \ForEach{$e=\hat{E}^{\dagger}{\hat{\vartheta}}$}{
 $\rho(e(t_n))\backsim{\mathcal{N}}({\mu_n},{\Sigma_n})$\\
 $\rho(e_n|t)\backsim{\mathcal{N}}({\hat{\mu_n}},{\hat{\Sigma_n}})$\\
 $\rho(e_p(t^*))\backsim{function}{(\E(e(t)),\D(e(t)))}$
  }
  }
\caption{Updated KS Framework}
\label{algorithm}
\While{$({p^{(i)}}\in{P_i})\longrightarrow{P_p}$}{
 $function (P_p) = (Inliers,Outliers)$\\
 $function (P_p,C_k) = (O)$\\
 $function (l,\psi) = SVM(O,model)$\\
 ${P_o}^c = centroid(\psi)$ \\
 ${e_O} = mapping({O_p})$
}
\ForEach{{$e^{t+1}_O$}$-${$e^{t}_O$}$\longrightarrow${$\Delta$$e_O$}}{$f_c$$\backsim${function ($O_p$, $P_c$)}\\
{$n_e$}$\backsim${function ($f_c$, {$\Delta$$e_O$})}\\
}
\end{algorithm}

Where, $M_{np}$ is the minimum number of points within the neighbour $N_\epsilon$ of candidate object's cluster under the radius of $\epsilon$, points A and B can be grouped together, if the euclidean distance between them is less than or equal to $\epsilon$ else discarded and $K_c$ represents the number of clusters for the localized candidate objects in the point cloud. 

The candidate objects, having segmented out from the point cloud in terms of clusters by Eq. \ref{eqt_5}, are further classified using a pre-trained SVM classifier that successfully recognizes $(\psi)$ and labels $(l)$ them. The pose of detected objects is then enumerated locally by normalizing respective clusters of the candidate objects to determine the 3D dimensions of their centroid (i.e, localizing the reference frame of object for pose information). Since, the estimated pose of objects is in physical coordinates and thus needs to be transformed into respective synergistic values to modulate the via/end points of the kernelized synergies in Eq. \ref{eqt_2}. Such a transformation is performed by using Eq. \ref{eqt_7}. 

\begin{equation}
    {{e_O}}={C_h}{\hat{E}^{\dagger}}{{{O_p}}} \int {{A^{\dagger}_{m}}}
\label{eqt_7}
\end{equation}

Where, $O_p$ is the pose of object in the Cartesian space, $C_h$ is the robot hand joint compliance matrix, $A_m$ is the motion transfer matrix and $\dagger$ represents the pseudo inverse of the respective variables. Finally, to execute the given task in the joint space of the robot hand, the necessary mapping is performed using $\vartheta = \hat{E}e + \theta_0$, with $\theta_0$ representing the initial configuration of the robot hand. 

\subsection{Force Feedback and Adaptation}

Robust grasping and fine manipulation of the detected candidate objects (from the previous subsection) require to regulate the force profile of the robot hand to avoid any possible damage or failure. Therefore, the general solution to the problem of robot hand (contact forces) balancing the grasped object is defined by Eq. \ref{eqt_4} \cite{c14}.

\begin{equation}
    {f_c}={G}^{\dagger}{\omega}+{\xi}{\Delta}{q_{ref}}
\label{eqt_4}
\end{equation}

Where, ${\delta}{q_{ref}}$ is the change in reference position of the robot hand joints, $\xi$ is the vector of internal forces generated during robot-object interaction, $\omega$ is the vector of external spatial forces and $G^{\dagger}$ is the pseudo inverse of the grasp matrix. Such a grasping problem when defined in the synergistic subspace can be formulated by Eq. \ref{eqt_11}, with ${\delta}{q_{ref}} = \hat{E}e$;

\begin{equation}
    {f_c}={G}^{\dagger}{\omega}+{\xi}{\hat{E}}{\Delta}{e}
\label{eqt_11}
\end{equation}
   
The contact forces $(f_c)$ applied by the robot hand on the grasped object are related to the current values $(I)$ of joints' motors by $f_c = {J_h}{K_m}I$, where $J_h$ is the robot hand Jacobian matrix and $K_m$ is the motor design constant. To control the motor currents, the tactile feedback from T40S chip sensors \cite{c15}, generating the 3D force information (mapped to the current values using deep ANN), is exploited, as shown in Fig. \ref{pose} (c). Further, to ensure the grasp stability and to avoid object slippage, the friction cone criteria defined by $\frac{F_{cz}}{\sqrt{{F_{cx}}^{2}+{F_{cy}}^{2}}}>\mu$, is imposed. Where, $\mu$ is the coefficient of friction and is tuned manually, depending upon the nature of task and the dynamics of system \cite{c16}. Note that all the coordinates (object poses, robot arm positions and orientations, and robot hand forces) are referred to the robot base frame using appropriate homogeneous transformation matrices.

Overall implementation and execution of updated kernelized synergies framework is summarized in Algorithm \ref{algorithm}. 

\section{Kernel Function Evaluation}

   \begin{figure*}[t]
      \centering
      \includegraphics[height=4.3cm, width=17.5cm]{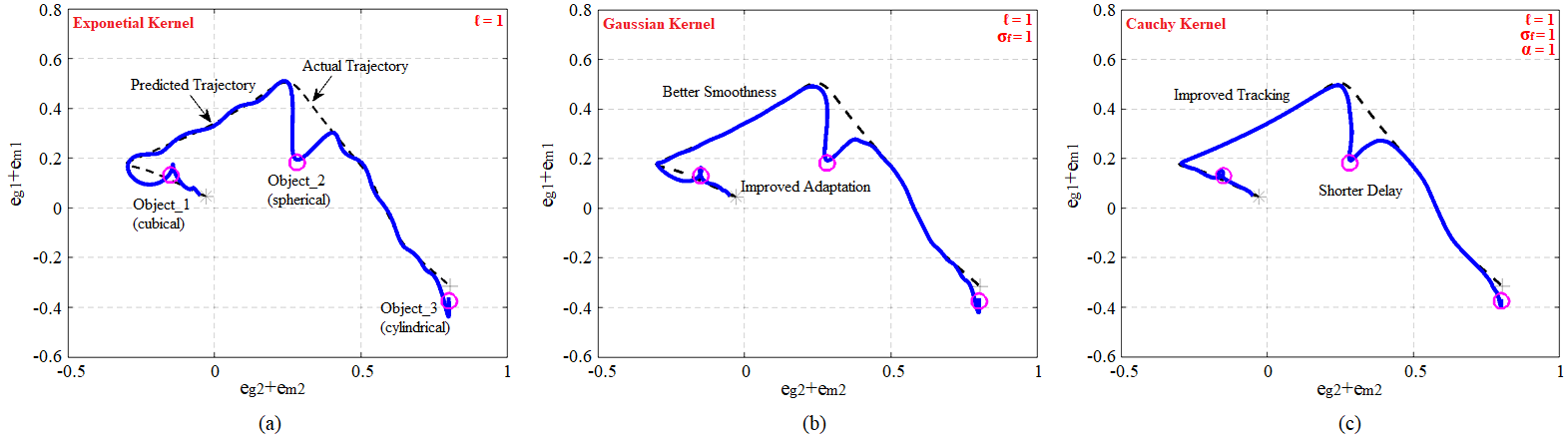}
      \caption{Performance evaluation of kernelized synergies framework against different kernel functions. (a), (b) and (c) represent the reference synergistic trajectory reproduced by using Exponential, Gaussian and Cauchy kernels for grasping of three different objects. The actual trajectory is generated by GMM-GMR model and the predicted trajectory is emulated by KMP using respective kernel functions.}
      \label{kernels}
      \vspace{-15pt}
   \end{figure*}

A kernel function $k(t_1,t_2)$ defined in Eq. \ref{eqt_2}, measures the similarity and correlation between two components of the synergistic subspace governing the robotic actions. The choice of a kernel function and its hyper-parameters do not only influence the synergistic trajectory distribution i.e, shape, smoothness and noise but also the characteristics of the kernelized synergistic subspace i.e, synergistic reusability and task generalization during model inference, as reported in Table \ref{grasp_quality}. For a kernel function to be valid, it is necessary and sufficient condition that the resulting kernel matrix $K$ in Eq. \ref{eqt_2} should be symmetric and positive semi-definite. Therefore, to investigate the performance and potential of the kernelized synergies framework, three valid kernel functions including; Exponential, Gaussian and Cauchy are considered using two evaluation metrics i.e, co-relation coefficient $(R)$ and root mean squared error $(rMSE)$, as quantified in Table \ref{grasp_qualities}. The $R$ measures a linear relationship between the actual and predicted values of the reference synergistic trajectory whereas the $rMSE$ evaluates the difference between two, according to Eqs. \ref{eqt_A} and \ref{eqt_B} respectively.             
\begin{equation}
    R=\frac{\sum_{i=1}^{M}(a_i-\hat{a}_{i})(p_i-\hat{p}_{i})}{\sqrt{\sum_{i=1}^{M}(a_i-\hat{a}_{i})\times{\sqrt{\sum_{i=1}^{M}(p_i-\hat{p}_{i})}}}}
\label{eqt_A}
\end{equation}

\begin{equation}
    rMSE=\sqrt{\frac{1}{M}\sum_{i=1}^{M}(a_i-p_i)^{2}}
\label{eqt_B}
\end{equation}

Where, $a_i$ and $p_i$ represent the actual and predicted values of the synergistic components, $\hat{a}_{i}$ and $\hat{p}_{i}$ are their respective average values and M is the total number of points in the synergistic trajectory distribution. 

\begin{table}[t]
\caption{Performance Evaluation of Kernelized Synergies Framework using different kernel functions against model characteristics.}
\label{grasp_quality}
\begin{center}
\begin{tabular}{P{4.5em}|P{4.5em}|P{5.5em}|P{4em}|P{4em}}
\hline
\textbf{Kernel} & \textbf{Synergistic Reusability} & \textbf{Task Reproducibility} & \textbf{Task Generalization} & \textbf{Task Execution} \\
\hline
\textit{Exponential} & {Yes} & {Oscillatory} & {Partially} & {Delayed} \\
\hline
\textit{Gaussian} & {Yes} & {Smooth} & {Fully} & {Reasonable} \\
\hline
\textit{Cauchy} & {Yes} & {Smoother} & {Fully} & {Improved} \\
\hline
\end{tabular}
\end{center}

\begin{footnotesize}
\noindent Synergistic Reusability: preserve the interaction properties of the subspace\\ 
\noindent Task Reproducibility: track and repeat the learned tasks during inference \\
\noindent Task Generalization: adapt to new shape and size of objects\\
\noindent Task Execution: complete tasks without deviations and delays\\
\end{footnotesize}
\end{table}

The exponential kernel defined by Eq. \ref{eqt_C}, is a better choice for approximating the non-differentiable functions and is thus unable to model all possible variations in the reference synergistic trajectory. This kernel is parameterized by the length scale $(l)$ and scale factor $(\sigma^2)$, regulating the duration and peak of the co-variance respectively.

\begin{equation}
    k_{exp}(t_1, t_2) = {\sigma}^{2}\exp(-\frac{|t_1-t_2|}{l})
\label{eqt_C}
\end{equation}
   
The parameter $l$ has exponential decay which governs the co-relation between the synergistic components and thus leads to a short-term memory effect. This effect preserves the probabilistic properties and the interaction characteristics of the computed synergistic subspace for synergistic reusability and task generalization. Whereas, the parameter $(\sigma^2)$ controls the repetition and execution of task during model inference. The reference trajectory reproduced by using exponential kernel for grasping of three different objects (i.e, cubical, spherical and cylindrical) is shown in Fig. \ref{kernels} (a). It can be seen that the predicted trajectory oscillates and poorly adapts to the dimensions of objects. It is due to the fact that the reverting force generated by $l$ try to pull the trajectory back to its mean position and thus causes temporally localized changes into the output with $R=0.6137$ and $rMSE=0.0418$, as reported in Table. \ref{grasp_qualities}.

To compensate oscillations into the trajectory and overcome the short-term memory effect, it is desirable to use a kernel that is infinitely differentiable i.e, Gaussian kernel. The Gaussian kernel, defined by Eq. \ref{eqt_D}, considers all possible variations in the reference trajectory and thus makes it smooth. Similar to exponential kernel, this kernel has also two hyper parameters i.e, $l$ and ${\sigma^2}$.     

\begin{equation}
    k_{gauss}(t_1, t_2) = {\sigma}^{2}\exp(-\frac{|t_1-t_2|^2}{2l^2})
\label{eqt_D}
\end{equation}

The reference synergistic trajectory generated by using Gaussian kernel for given three objects is shown in Fig. \ref{kernels} (b). It is evident that the predicted trajectory has however precisely followed the actual one with $R=0.6314$ and $rMSE=0.0378$, as defined in Table. \ref{grasp_qualities} as compared to Exponential kernel but yet has some periodic deviations and delays in tracking back the original path that affect the overall task execution.

To overcome the problem of periodic deviations and delays and also to maintain the temporal correlation between the synergistic components, the Cauchy kernel is exploited. This kernel is defined by Eq. \ref{eqt_E} and has similar characteristics as Gaussian kernel but uses Cauchy density instead, which introduces an additional parameter $\alpha$, called mixing coefficient to define the relative weights between different temporal correlations. The temporal correlations in this kernel have polynomial decay that leads to a long-term memory effect and a smoother and swift trajectory, as shown in Fig. \ref{kernels} (c). Hence, it gives an improved trajectory tracking and object adaptation with $R=0.6432$ and $rMSE=0.0362$ as compared to Exponential and Gaussian kernels in Table. \ref{grasp_qualities}.        

\begin{equation}
    k_{cauchy}(t_1, t_2) = {\sigma}^{2}(1+\frac{|t_1-t_2|^2}{2{\alpha}l^2})^{-\alpha}
\label{eqt_E}
\end{equation}

\section{RESULTS AND DISCUSSION}

\begin{table}[t]
\caption{Quantitative analysis of kernelized synergies with different kernels using statistical indices.}
\label{grasp_qualities}
\begin{center}
\begin{tabular}{P{3.5em}|P{4.5em}|P{5.5em}|P{4.5em}}
\hline
\textbf{Index} & \textbf{Exponential} & \textbf{Gaussian} & \textbf{Cauchy} \\
\hline
\textit{R} & {0.6137} & {0.6314} & {0.6432} \\
\hline
\textit{rMSE} & {0.0418} & {0.0378} & {0.0362} \\
\hline
\end{tabular}
\end{center}
\end{table}

To practically examine and evaluate the potential of upgraded kernelized synergies framework using Cauchy kernel, two tasks from daily life activities are considered i.e, (1) picking and placing an egg into the tray, (2) pouring tomato ketchup onto the food. The experimental setup consists of an anthropomorphic robot hand (Inspire Robotics), a redundant robot arm (Franka-Emika), a depth camera (Intel RealSense), tactile sensors (Touchence) and working environment (objects on the table), as shown in Fig. \ref{pose} (a). 

   \begin{figure*}[t]
      \centering
      \includegraphics[height=2.8cm, width=17.6cm]{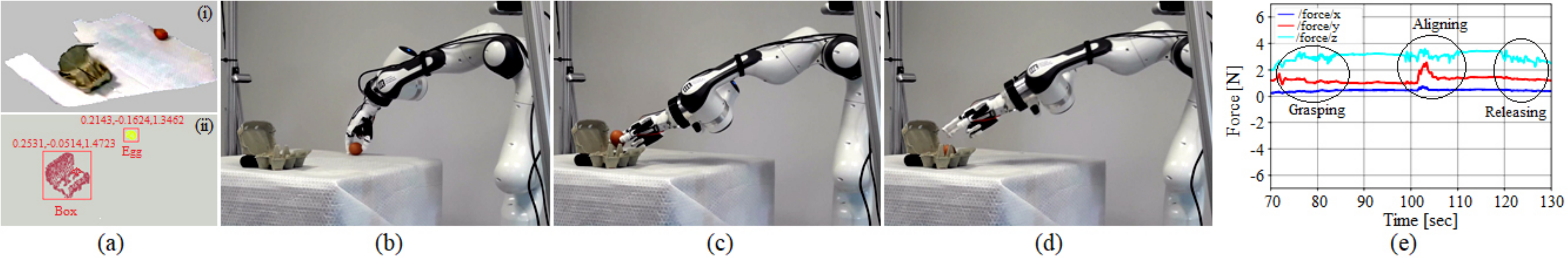}
      \caption{A Robot placing an egg into the tray, (a-i) is the raw point cloud of the egg-tray scene, (a-ii) shows the estimated poses of an egg and the desired chamber of the tray, in (b) the robot hand grasps the egg in its tripod posture, in (c) robot brings the egg over the tray, in (d) robot places the egg into the tray with precise orientation (aligning) and clearance (releasing), (e) represents the force feedback profile on grasping and manipulation of egg during the task execution.}
      \label{bulb}
      \vspace{-1pt}
   \end{figure*}
   
   \begin{figure*}[t]
      \centering
      \includegraphics[height=2.8cm, width=17.6cm]{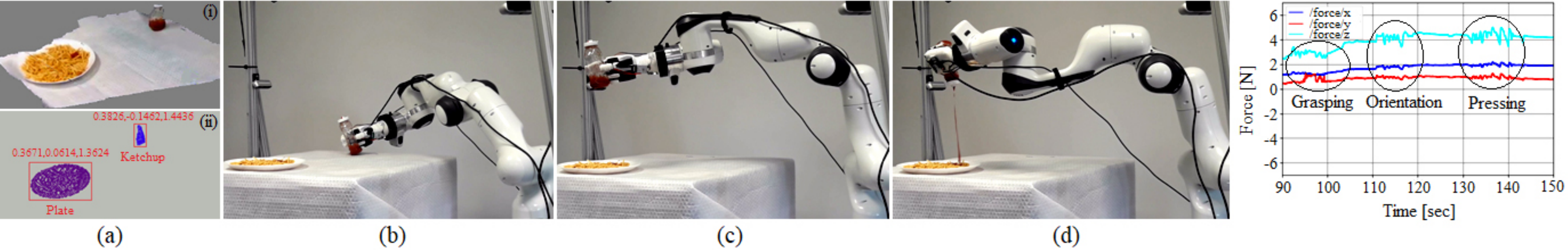}
      \caption{A Robot pouring tomato ketchup onto the food, (a-i) is the raw point cloud of the ketchup-food scene, (a-ii) shows the estimated poses of the ketchup bottle and the plate of food, in (b) the robot hand grips the bottle from middle in a tripod posture, in (c) robot brings the bottle over the plate and orients it $180^0$ anti-clockwise, in (d) robot presses the bottle to pour ketchup onto the food, (e) is the force feedback on desired interaction with the bottle during the task execution.\\}
      \label{lemon}
      \vspace{-15pt}
   \end{figure*}

For the first task, the robot hand exploits the rotation primitives of its synergistic subspace to properly place an egg into its tray (box). Firstly, the poses of egg and its corresponding tray on the table are estimated using proposed object detection pipeline according to Eqs. \ref{eqt_5} and \ref{eqt_7}, as shown in Fig. \ref{bulb} (a). This information tweaks the environmental descriptors of the kernelized synergies using Eq. \ref{eqt_2} and the updated values are used to grip the egg in a tripod posture at $e_{g1}=-0.163$, $e_{g2}=0.231$, as shown in Fig. \ref{bulb} (b). The mass of a given egg is about $68$ grams and is placed in quadrature to robot hand z-axis to ease its grasping but may also assume any other position, thanks to the robustness of updated framework against the perceptual feedback. The egg is brought over the tray using robot arm movements in Fig. \ref{bulb} (c) and is inserted into the desired chamber with minor clockwise orientation at $e_{m1}=-0.154$ to $-0.09$, $e_{m2}=0.242$ to $0.273$ such that the axis of egg gets aligned with the chamber in Fig. \ref{bulb} (d). In this task the force profile of the robot hand is regulated approximately from $2.38$ to $3.16$ $N/sec$ using tactile feedback according to Eq. \ref{eqt_11}, as shown in Fig. \ref{bulb} (e). This does not only tackle the slippage but the damage of egg as well, owing to its brittle nature. Moreover, for this example, the value of $\mu$ is $0.64$, which is determined empirically during the trial phase.
   
 In the second task, the robot hand squeezes tomato ketchup bottle using translation primitives of its synergistic subspace. The mass of a considered ketchup bottle is around $117$ grams and its cap is left open intentionally to simplify the pouring action. Having determined the poses of bottle and food (plate) in Fig. \ref{lemon} (a), the robot hand grasps the bottle from middle in a tripod posture at $e_{g1}=0.144$, $e_{g2}=0.283$ in Fig. \ref{lemon} (b). The bottle is brought over the plate and rotated anti-clockwise such that its opening (z-axis of the bottle) is facing downwards using robot arm movements in Fig. \ref{lemon} (c). The robot hand stretches its index and middle fingers against the thumb at $e_{m1}=0.152$ to $0.311$, $e_{m2}=0.293$ to $0.486$ so that the bottle is pressed and tomato ketchup is poured onto the food in Fig. \ref{lemon} (d). During this task, the interaction force between the robot hand and bottle is modulated from $2.38$ to $4.26$ $N/sec$, according to Eq. \ref{eqt_11}, as reported in Fig. \ref{lemon} (e). The value of $\mu$ for this example is $0.71$.

\section{CONCLUSIONS}

In this research, the previously developed kernelized synergies framework has been augmented with visuo-tactile perception for autonomous interaction and run-time adaptation. The visual information in-terms of point cloud has been processed using RANSAC algorithm together with SVM classifier for semantic segmentation, classification and pose estimation. For flexible robot-object coordination, the tactile feedback has been used to modulate the current values of the robot hand joints' motors to regulate the force profile of the kernelized synergies accordingly. Moreover, the performance and effectiveness of kernelized synergies have been tested against the synergistic re-usability and task reproducibility, generation and execution using three different valid kernel functions i.e, Exponential, Gaussian and Cauchy. Hence, it has been found that the Cauchy kernel, having long-range memory effect and polynominal temporal co-relations, outperforms the other two and is also robust against perceptual uncertainty and feedback latency. Finally, the tasks performed using anthropomorphic robot arm-hand system for fine grasping and manipulation of non-rigid objects (egg and ketchup bottle) has confirmed the improved accuracy and potential of updated kernelized synergies framework.

Considering the adaptivity, priority characteristics and movement primitive nature of the kernelized synergies, the updated framework will be evaluated for whole-body motion planning of non-holonomic manipulation tasks in the field agricultural robotics, i.e, extending the idea discussed in \cite{c17}.   

\bibliographystyle{IEEEtran}

\end{document}